\documentclass [10pt,conference,letterpaper] {IEEEtran}
\pdfoutput=1

\usepackage{cite}
\usepackage{graphicx} %use graph format
\usepackage{epstopdf}
\usepackage[ruled]{algorithm2e}
\usepackage{algpseudocode}  
\usepackage{amsmath}  
\usepackage{xcolor}

\usepackage{enumerate}
\usepackage{subfigure}
\usepackage{url}
\usepackage{tcolorbox}
\usepackage{url}
\usepackage{multirow}
\usepackage{tabularx}

\usepackage{amsthm,amsmath,amssymb, newtxmath} 
\usepackage{mathrsfs}
\usepackage{bbm}

\ifCLASSINFOpdf 
\else
\fi

\begin{document}

\newcommand{\ModelName}{\texttt{TrafficGPT}}

\title{{\huge$\mathbbm{Traffic}$}$\mathbb{GPT}$:
Breaking the Token Barrier for  
Efficient Long Traffic
Analysis and Generation}
% with Linear Complexity Transformers}

\IEEEoverridecommandlockouts

\author{\IEEEauthorblockN{
        Jian Qu\IEEEauthorrefmark{2}\IEEEauthorrefmark{4},
        Xiaobo Ma\IEEEauthorrefmark{2}\IEEEauthorrefmark{4}, 
        Jianfeng Li\IEEEauthorrefmark{2}\IEEEauthorrefmark{4} 
	\IEEEauthorblockA{\IEEEauthorrefmark{2}MOE Key Lab for Intelligent Networks and Network Security, Xi'an Jiaotong University,  Xi'an, China}
			\IEEEauthorblockA{\IEEEauthorrefmark{4}Faculty of Electronic and Information Engineering, Xi'an Jiaotong University, Xi'an, China}
	}	
}

\maketitle

\begin{abstract}

Over the years, network traffic analysis and generation have advanced significantly. From traditional statistical methods, the field has progressed to sophisticated deep learning techniques. This progress has improved the ability to detect complex patterns and security threats, as well as to test and optimize network performance.  However, obstacles persist, such as the dependence on labeled data for analysis and the difficulty of generating traffic samples that follow realistic patterns.  Pre-trained deep neural networks have emerged as powerful tools to resolve these issues, offering improved performance by learning robust data representations from large unlabeled datasets. Despite their benefits, existing pre-trained models face challenges like token length limitation, which restricts their usefulness in comprehensive traffic analysis and realistic traffic generation.
To address these challenges, we introduce \ModelName, a deep learning model that 
can tackle complex challenges related to long flow classification and generation tasks. This model 
uses generative pre-training with the linear attention mechanism, which allows for a substantially increased capacity of up to 12,032 tokens from the previous limit of only 512 tokens. 
\ModelName\  demonstrates superior performance in classification tasks, reaching state-of-the-art levels. In generation tasks, it closely resembles real traffic flows,  with low JS divergence and an F1 score close to  0.5  (representing a random guess) in discriminating generated data. 
These advancements hold promise for future applications in both traffic flow classification and generation tasks.

\end{abstract}

\IEEEpeerreviewmaketitle

\section{{Introduction}}

The analysis and generation of network traffic have long been two critical tasks. Network traffic analysis can be utilized to identify patterns, detect security threats, and optimize network performance among other applications\cite{papadogiannaki2021survey,abbasi2021deep, javaheri2023fuzzy}. Meanwhile, network traffic generation can be used to simulate various scenarios for testing network infrastructure, validating security measures, and training machine learning models to recognize and respond to different network behaviors\cite{yin2022practical, adeleke2022network}.

% 历史 + 现状 1
Network traffic analysis has made significant strides in recent years, transitioning from traditional statistical methods to more advanced deep learning techniques. Early approaches heavily relied on manually crafted features, which limited their ability to capture complex patterns in raw traffic data\cite{hayes2016k,al2016adaptive, taylor2017robust}. However, the advent of deep learning methods has revolutionized this field by enabling automatic extraction of intricate patterns, leading to remarkable performance improvements\cite{rimmer2017automated,liu2020attention,luo2022transformer,song20232,qu2023input}. Despite these advancements, a critical obstacle remains, i.e.,  the dependency on labeled training data. The quantity and distribution of labeled data greatly influence the effectiveness and robustness of deep learning models,  leading to biases and poor generalization in real-world scenarios.

% 历史 + 现状 2
On the other hand, significant progress has been made in generating network traffic, especially with the emergence of software-defined networking and network function virtualization. Research in this field has led to the development of experimental environments that resemble actual networks in terms of node variety and network topology \cite{adeleke2022network}. However, generating diverse and realistic traffic patterns continues to be a major challenge. Despite the increased accessibility of experimental setups, creating traffic that accurately reflects real-world scenarios remains a difficult task.

% 预训练模型的现状
In recent times, pre-trained deep neural networks have emerged as leading methodologies for both network traffic analysis and generation tasks. One such model, ET-BERT\cite{lin2022bert}, uses the BERT architecture and  has showcased superior performance compared to models without pre-training across various traffic classification tasks. Another model, Lens\cite{wang2024lens},   employs the T5 architecture for pre-training and has achieved state-of-the-art results in generating packet header fields. By leveraging large  amounts of unlabeled data, pre-training-based approaches adeptly learn robust representations. Subsequently, these representations can be seamlessly applied to downstream tasks through fine-tuning with limited labeled data, exemplifying pre-training's versatility and efficacy in network analysis and generation.

% 面临的挑战
While pre-trained models have many benefits, they encounter two primary challenges. Firstly, the tokenization process in these  models needs refinement. Existing methods of tokenization in pre-trained models have shortcomings, as they struggle to accurately reconstruct pcap files from the token lists generated by the model. This limitation hinders their practical usefulness. Secondly, pre-trained models have a significant constraint on token length. Most pre-trained models used for traffic analysis are restricted to a maximum of 512 tokens. This limit is insufficient for realistic traffic analysis.   This issue becomes even more pronounced in traffic generation, where the token count for a single packet can exceed 512, making it difficult to generate real-world traffic samples.

% 我们的方案
To address these challenges, we propose \ModelName, a deep-learning model that leverages generative pre-training with the linear attention mechanism. Starting with the tokenization issue, we develop a reversible token representation method. This approach allows for the direct generation of pcap files from token lists, effectively solving the problem of accurately reconstructing traffic flows from the model's output. Furthermore, to overcome the token length limitation, we implement a linear attention mechanism in place of the traditional quadratic self-attention mechanism found in Transformer\cite{vaswani2017attention}. This modification significantly increases the model's capacity, supporting a maximum token length of 12,032. Together, these enhancements greatly enhance the model's capabilities in both traffic analysis and generation.

% 贡献总结
Our major contributions are summarized as follows.

\begin{itemize}
	
\item {We introduce \ModelName, a deep-learning model using generative pre-training. It utilizes a linear attention mechanism to replace the traditional quadratic self-attention mechanism in Transformer, enabling a token scope of up to 12,032 tokens, making it suitable for both flow classification and generation tasks.}
	
\item {We develop  a reversible token representation method, which  enables bidirectional mapping between pcap files and token representation.
This approach facilitates the direct generation of pcap files from token lists, effectively addressing the challenge of accurately reconstructing traffic flows from the model's output.}

\item {Our model performs exceptionally well in classification experiments, achieving state-of-the-art results with an average improvement of 2\% in Macro F1-Score across various datasets. 
In the generation evaluation, our model demonstrates its ability to generate traffic flows similar to real ones, with an average JS divergence of 0.1605 for packet headers and 0.2396 for flow features.  
Moreover, the F1 score for discriminating our generated flows is 0.6683, very close to  0.5 (representing a random guess), indicating that our generated flows are highly realistic and difficult to distinguish from actual ones.}

\end{itemize}

%每节干了什么
\noindent
\textbf{Roadmap.} Sec. \ref{sec:Related Work} introduces related work.
Sec. \ref{sec:System Design} elaborates system design and Sec. \ref{sec:Evaluation} evaluates it.
Sec. \ref{sec:Discussion} discusses the limitations in our work and promising directions for future research.
We conclude in Sec. \ref{sec:Conclusion}.

\section{{Related Work}}
\label{sec:Related Work}

This section provides an overview of the existing literature related to our work, focusing on three main areas: traffic classification, network traffic generation, and advancements in Transformer architectures for handling long sequences efficiently. Each of these areas contributes to the foundation upon which our proposed model is built, addressing the challenges and limitations encountered in current methodologies.

\noindent
\textbf{Traffic Classification.}
% 预训练模型
There are several papers on the large-scale pre-training of models in the field of traffic\cite{lin2022bert, meng2023netgpt, zhao2023yet, guthula2023netfound,he2020pert,wang2024lens}.
He \textit{et al.} pretrained a transformer on the payload of encrypted packets\cite{he2020pert}.
Lin \textit{et al.} extracted bursts from the traffic and used the burst bytes as the pretext for pre-training the BERT model, naming it ET-BERT\cite{lin2022bert}.
Meng \textit{et al.} proposed a generative pre-trained Transformer and used the first three packets with a maximum token size of 512 for training and testing. The results outperformed ET-BERT in several tasks\cite{meng2023netgpt}.
Zhao \textit{et al.}  introduced a masked autoencoder-based model for traffic classification, converting the initial five packets of each flow into images and subsequently employing pre-training based on the Transformer\cite{zhao2023yet}.
Guthula \textit{et al.} put forth a hierarchical Transformer architecture for flow modeling, incorporating a packet-burst-flow structure\cite{guthula2023netfound}.
Wang \textit{et al.} introduced Lens, a model that leverages the T5 architecture to learn representations from large-scale data\cite{wang2024lens}.

% 没有预训练的模型
Prior to the advent of pre-trained models, researchers typically relied on small-scale datasets for model training and testing.
Hayes \textit{et al.}  proposed a system for website fingerprinting on Tor, utilizing random forests to extract fingerprints\cite{hayes2016k}.
Yan \textit{et al.} examined the situation of keyword-based searching fingerprinting through the development of a hand-crafted feature set\cite{yan2020fingerprinting}.
Rimmer \textit{et al.} introduced a website fingerprinting attack over Tor by comparing multiple neural network structures\cite{rimmer2017automated}.
Liu \textit{et al.} examined the use of an attention-based bidirectional gated recurrent unit neural network for the identification of HTTPS web services\cite{liu2020attention}.
Holland \textit{et al.} integrated nPrint with automated machine learning, streamlining the workflow for traffic classification\cite{holland2021new}.
Luo \textit{et al.} developed a Transformer-based IoT device-type identification method\cite{luo2022transformer}.
Lin \textit{et al.} devised an adaptive balancing training method to address dataset imbalances and employed multi-level features for detecting malicious traffic\cite{lin2022mffusion}.
Li \textit{et al.}  achieved open-world Android app user action identification via synthesizing traffic and binary analysis\cite{li2022foap}.
Song \textit{et al.} proposed an incremental and interpretable recurrent neural network model for encrypted traffic classification\cite{song20232}.
Qu \textit{et al.}  designed a hierarchical deep learning model capable of integrating multiple flows of information\cite{qu2023input}.
Guan \textit{et al.} leveraged federated learning for encrypted traffic classification\cite{guan2023personalized}.
Xie \textit{et al.} devised data augmentation techniques tailored for TCP traffic, leveraging BYOL\cite{grill2020bootstrap} for self-supervised learning of robust features\cite{xie2023rosetta}.

\noindent
\textbf{Network Traffic Generation.}
Adeleke \textit{et al.} comprehensively analyzed the traffic generation tools used by researchers over the past decade, including 92 different tools such as application layer generators, traffic replay tools, model-based traffic generators, and more\cite{adeleke2022network}.
Ring \textit{et al.} utilized Generative Adversarial Network (GAN) techniques to generate flow-based characteristics\cite{ring2019flow}.
Cheng \textit{et al.} utilized a convolutional neural network-based GAN to conduct the generation of IP packets\cite{cheng2019pac}.
Manocchio \textit{et al.} mitigated the issue of model collapse by incorporating the concept of Manifold Guided Generative Adversarial Networks in the synthetic generation of network flow\cite{manocchio2021flowgan}.
Fan  \textit{et al.}  integrated the concept of differential privacy into GANs to generate features of flows, aiming to achieve secure sharing of network data\cite{fan2021dpnet}.
Zolbayar \textit{et al.} utilized GAN to generate traffic features and investigated their adversarial impact on certain machine learning classifier-based methods in whitebox, blackbox, and restricted-blackbox threat models \cite{zolbayar2022generating}.
Hui \textit{et al.} introduced a knowledge-enhanced GAN framework for large-scale IoT traffic generation, addressing the limitations of existing IoT synthetic data methods\cite{hui2022knowledge}.
Yin \textit{et al.} utilized a time-series GAN to generate packet header fields in the flows\cite{yin2022practical}.
Du \textit{et al.}   adopted dynamic word embedding and long short-term memory networks to generate the communication patterns between IP addresses and ports\cite{du2023dbwe}.
Kim  \textit{et al.} employed GAN to simulate and generate the spectral data in the context of 5G networks\cite{kim2022design}.
Kholgh \textit{et al.} fine-tuned GPT-3 to generate ICMP and DNS packets\cite{kholgh2023pac}.

\noindent
\textbf{Efficient Transformers.}
Keles \textit{et al.} prove that the time complexity of self-attention is necessarily quadratic in the input length under the Strong Exponential Time Hypothesis\cite{keles2023computational}.
This implies that currently we can only resort to approximate algorithms to reduce the complexity of self-attention.
Guo \textit{et al.}\cite{guo2019star}, Beltagy \textit{et al.}\cite{beltagy2020longformer}, Zaheer \textit{et al.}\cite{zaheer2020big} and  Roy \textit{et al.}\cite{roy2021efficient}  respectively employ different types of sparse self-attention to reduce computational overhead.
Katharopoulos \textit{et al.} replaced the dot product with a simple feature map, achieving linear complexity in the Transformer\cite{katharopoulos2020transformers}.
Lee \textit{et al.}\cite{lee2019set}, Wang \textit{et al.}\cite{wang2020linformer} and Zhang \textit{et al.}\cite{zhang2021poolingformer} compressed key–value memory in different ways and achieved linear complexity.
Guo \textit{et al.}  developed low-rank attention and band attention to parameterize the self-attention mechanism\cite{guo2019low}.
Fan \textit{et al.} further used the low-rank decomposed self-attention to achieve the linear complexity\cite{fan2021lighter}.
Kitaev \textit{et al.}  used Locality-Sensitive Hashing and reversible residual network to reduce computational cost\cite{kitaev2020reformer}.
Peng \textit{et al.}  proposed a new attention mechanism reformulation that results in linear attention\cite{peng2023rwkv}.
Sun \textit{et al.} proposed the Retentive Network, demonstrating its performance to be comparable to that of a Transformer of similar size in language modeling. 
They highlighted its advantages, including training parallelism, cost-effective deployment, and efficient inference \cite{sun2023retentive}.

\section{{System Design}}
\label{sec:System Design}

% 在本节中，我们将详细介绍ABC模型的设计，并重点指出我们的模型与ETbert和NetGPT的不同之处。

We present \ModelName, a deep-learning model that leverages generative pre-training with the linear attention mechanism. Our approach integrates fundamental principles from ET-BERT\cite{lin2022bert} and NetGPT\cite{meng2023netgpt} while introducing refinements to optimize token representation and enhance the neural network architecture. Our model is tailored to effectively handle long token sequences, improving the generation and classification of network traffic.

\subsection{Model Architecture}
\begin{figure*}[tbp]
    \centering
    \includegraphics[width=0.95\linewidth]{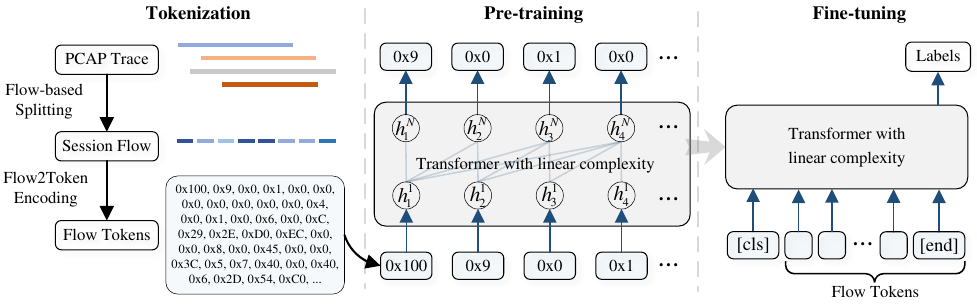}
    \caption{The framework of \ModelName.}
    \label{fig:framework}
\end{figure*}

The primary objective of \ModelName \  is to learn and represent universal features, focusing on network flows as the basic unit.
This model will be tested in scenarios involving traffic generation and classification.
Its methodology encompasses a series of steps, as depicted in Figure \ref{fig:framework}. Initially, it converts network flows into token representations, followed by extensive pre-training on large-scale traffic data. The final phase involves testing the model's proficiency in understanding traffic tasks, including traffic generation and classification.

During the token representation phase, each network flow is meticulously mapped into a list of tokens. This bijective process ensures that every token list can be accurately converted back into its original flow, thereby maintaining the integrity of the extracted information. In the pre-training stage, the model engages in self-supervised learning using unlabeled flows, employing an autoregressive approach to develop a comprehensive feature representation of network traffic.
For the final application stages, strategies for traffic generation and fine-tuning approaches for traffic classification are thoughtfully formulated.

Architecturally, \ModelName \  is grounded in a linear attention mechanism\cite{katharopoulos2020transformers}, enhanced by integrating local attention strategies\cite{xiong2021simple} and the reversible network in Reformer\cite{kitaev2020reformer}, effectively optimizing memory usage. 
The inclusion of token shift\cite{peng2023rwkv} is a strategic choice to expedite the model's convergence. 
The mechanisms are detailed in Appendix \ref{sec: principle}.
The model is characterized by a hidden layer dimension of 512, encompassing 12 attention heads and spanning a depth of 24 layers. 
For a comprehensive breakdown of additional parameters, kindly consult Section \ref{subsec:Settings}.

\subsection{Tokenization}
In the tokenization stage, we optimized the tokenization processes of both ET-BERT\cite{lin2022bert} and NetGPT\cite{meng2023netgpt}, achieving a more seamless overall workflow. 
One key innovation is the integration of time information into tokens. It empowers \ModelName \ to generate timestamp intervals for pcap files. By incorporating temporal data into our tokenization strategy, we have enabled a more native and comprehensive representation of the information contained in pcap files.

\begin{figure}[tbp]
    \centering
    \includegraphics[width=1\linewidth]{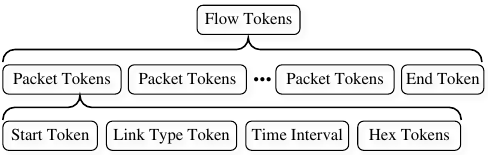}
    \caption{The structure of flow tokens.}
    \label{fig:encoding}
\end{figure}

Firstly, we segment the pcap files within the dataset into distinct flows. A flow is precisely defined by a quintuple, representing a sequence of packets sharing identical source addresses, destination addresses, source ports, destination ports, and protocols. Following this segmentation, we proceed to tokenize each flow.

Figure \ref{fig:encoding} elucidates the token composition for a singular flow. Within this framework, each flow comprises tokens corresponding to multiple packets, culminating in an end token denoting termination. The tokens assigned to each packet can be dissected into four components:

\noindent
\textbf{Packet Start Token.} This pivotal token signifies the commencement of a packet, playing a fundamental role in clearly defining the boundaries between individual packets.

\noindent
\textbf{Link Type Token.} This token denotes the specific link layer protocol in use, discerning between protocols like Ethernet or Linux cooked mode. Its critical importance in pcap generation stems from the distinct formats inherent in different link layer protocols.

\noindent
\textbf{Time Interval Tokens.} These tokens indicate the time interval between the current packet and the preceding one. 
In the case of the initial data packet, we establish its time interval as 0.
We transform the timestamp into exponential form, representing it with 8 bytes, where each byte functions as a distinct token.
%Each packet encapsulates a timestamp with microsecond precision, utilizing 8 bytes. 
%To enhance token efficiency, 

\noindent
\textbf{Hex Tokens.} These tokens encapsulate all pertinent information for each packet, encompassing values from both the packet header and payload. Considering the presence of both encrypted and non-encrypted data within packets, the decision to convert all bytes into hexadecimal tokens provides a universal representation.

\subsection{Pre-training}
To attain efficient multitasking performance, especially in generative tasks, we employ a pre-training approach akin to the auto-regressive method utilized in GPT-2. 
This methodology, characterized by the incremental generation of sequences, empowers the model to acquire nuanced representations of context, leading to improved generalization across diverse tasks.

% In this process, the model progressively predicts the next vocabulary token, utilizing previously generated content as context, thereby gradually enhancing its language understanding and generation capabilities. This auto-regressive method excels in generative tasks, enabling the model to produce text with contextual consistency and semantic coherence.

In this process, the model incrementally predicts the subsequent vocabulary token, leveraging previously generated content as context. This approach enables the model to refine its language understanding and generation capabilities. The auto-regressive method excels particularly in generative tasks, allowing the model to produce text with contextual coherence and semantic consistency.

Specifically, in the context of a given input traffic sequence denoted as $X = [x_1, x_2, ..., x_T]$, the model is trained to predict the probability distribution of the next token based on the preceding tokens in the sequence. This probability distribution is typically determined using a softmax activation function:
\begin{equation}
P(x_t | x_{1:t-1}) = \text{softmax}(f(x_{1:t-1}; \theta)).
\label{eq:autoreg} 
\end{equation}
Here, $x_t$ is the target token at position $t$, and $x_{1:t-1}$ represents the sequence of tokens from position 1 to $t-1$. The function $f$ embodies the model's parameters denoted as $\theta$.

For the training process, the auto-regressive pre-training employs cross-entropy loss, expressed as:
\begin{equation}
\text{Loss} = -\sum_{i=1}^{V} y_{t,i} \cdot \log(P(x_{t = i} | x_{1:t-1})),
\label{eq:autoregloss}
\end{equation}
In this equation, $V$ represents the vocabulary size, and $y_{t,i}$ is the one-hot encoded ground truth corresponding to the target token. This loss function quantifies the disparity between the predicted probability distribution and the actual distribution, guiding the model towards optimal token prediction.

\subsection{Traffic Generation}

In the traffic generation tasks, we kickstart the process by manually providing a straightforward start token or a predefined set of initial tokens. These initial tokens serve as the foundation for the generation process. They are then input into the pre-trained model, prompting it to predict the subsequent token sequentially until a termination token is reached.

It's important to highlight that proficiently trained models can generate sequences comprising a huge number of tokens. 
The length of these generated sequences can easily surpass the maximum window length established during the training phase, a limitation typically dictated by the available GPU memory.
This phenomenon is analogous to creating tokens through a sliding window. Even though the maximum token length set confines the model's perspective during training, it consistently generates tokens based on the preceding tokens within the window of its perspective. 
This continuous generation process enables the model to produce coherent and contextually relevant sequences.

We use Top-k sampling to enhance the quality and diversity of generated sequences.
Top-k sampling is a probabilistic method employed during the generation phase to select the next token from the model's probability distribution. Instead of choosing the token with the highest probability outright, Top-k sampling involves sampling from the top $k$ tokens with the highest probabilities, where $k$ is a predefined hyperparameter.
By restricting the potential choices to a smaller set of high-probability tokens, we prioritize the model's most confident predictions. This helps mitigate the risk of introducing irrelevant or nonsensical tokens into the sequence, contributing to more contextually relevant and coherent outputs.

The subsequent phase in the workflow involves translating the generated token sequence into a format suitable for representing network traffic data, such as a pcap (Packet Capture) file. This translation process is essentially the inverse of tokenization, whereby each token in the sequence is utilized to construct a corresponding segment of the network traffic data.

The process initiates with identifying and extracting individual data packets from the token list. This is achieved by leveraging a designated packet start token, which serves as a delimiter to delineate the boundaries of each packet within the sequence. Subsequently, parsing operations extract pertinent information from the tokens, such as the link type and hexadecimal representations.

It's crucial to note that, despite the model's proficiency, there's a slight probability of generating "illegal packets." These problems might occur when the model generates data packets that cannot be properly parsed due to protocol inconsistencies, such as undefined packet header fields or lengths exceeding protocol-defined limits. To mitigate the impact of such irregularities, a straightforward strategy is employed: any packet deemed "illegal" is discarded, and the generation process restarts from the beginning of the preceding packet using the start token. This iterative approach ensures the production of a coherent and protocol-compliant network traffic data sequence.

\subsection{Fine-tuning}
Fine-tuning involves adjusting the parameters of a pre-existing, pre-trained model to better suit the specific requirements of a particular task or domain. In this paper, we align our approach with methodologies seen in ET-BERT\cite{lin2022bert} and NetGPT\cite{meng2023netgpt}, focusing specifically on a classification task designed to categorize network flows.

Our model employs a straightforward yet effective method for fine-tuning. As a first step, we introduce a [cls] token at the beginning of the flow's token list. This addition signals the model that it is about to undertake a classification task. Subsequently, both the [cls] token and the flow's tokens are fed into the model. This allows the model to utilize the next output as the label for classification, as depicted in Figure \ref{fig:framework}.

The model we have developed can handle a total of 260 different tokens. Consequently, using a single token enables the classification of up to 260 distinct classes. In scenarios where the number of classes exceeds 260, we can expand this capacity by employing multiple tokens. For instance, using two tokens can facilitate labeling for as many as $260^2$ classes.

\section{{Evaluation}}
\label{sec:Evaluation}
We systematically assess the performance of \ModelName\  across various metrics and comparison analyses. 
We begin by outlining the settings, which include detailed dataset descriptions, data preprocessing methods, and hyper-parameter setups, laying the groundwork for an in-depth evaluation. Following this, we explore both classification and generation evaluations in detail. These analyses compare our models against existing methodologies in tasks such as traffic flow classification and generation.

\subsection{Settings}
\label{subsec:Settings}
\noindent
\textbf{Datasets.}
We have assembled a comprehensive compilation of five publicly accessible datasets, totaling an extensive 189 gigabytes in size. The included datasets are ISCXTor2016 \cite{lashkari2017characterization}, USTCTFC2016 \cite{USTCTFC2016}, ISCXVPN2016 \cite{draper2016characterization}, DoHBrw2020 \cite{montazerishatoori2020detection}, and CICIoT2022 \cite{dadkhah2022towards}, covering a diverse range of network traffic types operating within the TCP/IP framework. This encompasses terminal user internet activity, Virtual Private Network (VPN) traffic, Tor network traffic, and Internet of Things (IoT) communication. Notably, some datasets provide both traffic feature files and labels alongside pcap or pcapng files. Our exclusive focus in the analysis is on the raw packets, and any supplementary data has been disregarded for our study.

\noindent
\textbf{Data Pre-processing.}
In the data preprocessing phase, we start by categorizing traffic into flows using the five-tuple approach, considering source IP address, destination IP, source port, destination port, and protocol. 
This allows us to isolate and extract specific flows within the network. 
For traffic that doesn't fall under the TCP or UDP categories, such as Address Resolution Protocol (ARP) and Dynamic Host Configuration Protocol (DHCP) packets, we adopt a method akin to ET-BERT\cite{lin2022bert}, simply discarding them as they are irrelevant to the particular content being transmitted.
After that, 99\% of the flows is allocated for the pretraining process, while the remaining 1\% is set aside for testing.

\noindent
\textbf{Hyper-parameters.}
\ModelName\  is specified with 260 tokens, a feed-forward dimensional of 512, 12 attention heads, and a depth of 24 layers, catering to sequences up to 3,072 (3k) and 12,032 (12k) tokens in length. It incorporates a dropout rate of 0.1 for feed-forward layers, self-attention layers, and post-attention mechanisms to mitigate overfitting. 
The model utilizes an embedding dimension of 256 and a head dimension of 256. 
It is also complemented by 8 local attention heads and a local window of 256, enhancing the model's focus on relevant local sequence segments. The architecture is made reversible, drawing inspiration from the Reformer\cite{kitaev2020reformer}, to improve memory efficiency further. A GLU variant is used for enhanced non-linearity. Additionally, token shifting is applied to improve convergence. The training regimen involves a learning rate of $1\times 10^{-4}$, a batch size of 4, and spans 750,000 steps.

\subsection{Classification Evaluation}
\label{subsection:cls}
\begin{table*}[htbp]
    \footnotesize
    \caption{Comparison of Traffic Classification Macro F1-Scores.}
    \vspace{-1pt}
    \label{tab:fine-tune} 
    \centering
    \begin{tabular}{|>{\centering\arraybackslash}m{2.2cm}|>{\centering\arraybackslash}m{1.3cm}|>{\centering\arraybackslash}m{1.3cm}|>{\centering\arraybackslash}m{1.3cm}|>{\centering\arraybackslash}m{1.3cm}|>{\centering\arraybackslash}m{1.3cm}|>{\centering\arraybackslash}m{1.3cm}|>{\centering\arraybackslash}m{1.3cm}|>{\centering\arraybackslash}m{1.3cm}|}
        \hline
          \multirow{2}{*}{\textbf{Method}} & \multicolumn{2}{|c|}{{Cross-Platform(iOS) }} & \multicolumn{2}{|c|}{{Cross-Platform(Android)}} & \multicolumn{2}{|c|}{{ISCX-VPN-App}} & \multicolumn{2}{|c|}{{USTC-TFC}}  \\
        \cline{2-9}
           & AC & F1 & AC & F1  & AC & F1 & AC & F1\\
        \hline
        PERT\cite{he2020pert}&0.9789&0.9584&0.9772& 0.8550&N/A&N/A&N/A&N/A\\
        ET-BERT\cite{lin2022bert}&0.9844&0.9643&\textbf{0.9865}&0.9246 &0.9206& 0.4314&0.9524& 0.6986\\
        NetGPT\cite{meng2023netgpt}&N/A&N/A&N/A&N/A&0.9683&0.8056&0.9563&0.9463\\
        YaTC\cite{zhao2023yet}&0.9842&0.9644&0.9816&0.9217&0.9908&0.9860&0.8071&0.7452\\

        Lens\cite{wang2024lens}&0.9189&0.9143&0.9063&0.8981&0.9984 &0.9958&\textbf{0.9940}&\textbf{0.9937}\\
        \hline
        \ModelName(3k)&\textbf{0.9844}&0.9829&0.9540&0.9483 &0.9912 &0.9912 &{0.9856} &{0.9854}\\
        \ModelName(12k)&0.9839&\textbf{0.9863}&0.9444&\textbf{0.9498}&\textbf{1.0000}&\textbf{1.0000}&0.9900&0.9877\\
        \hline
    \end{tabular}
    \vspace{-1pt}
\end{table*}

We conduct experiments to evaluate the performance of flow classification tasks. We introduce two models, named \ModelName(3k) and \ModelName(12k), each pre-trained using distinct maximum token lengths of 3k and
12k, respectively. 

For the Cross-Platform datasets, encompassing both iOS and Android, we adopted the preprocessing approach detailed in ET-BERT\cite{lin2022bert}. This preprocessing approach involved removing flow files smaller than 5KB and discarding classes with insufficient data. As a result, the dataset for Cross-Platform (iOS) comprised 196 labels, while Cross-Platform (Android) contained 215 labels.

Regarding the ISCX-VPN-App and USTC-TFC datasets, our preprocessing was aligned with the methodology used in NetGPT\cite{meng2023netgpt}. In this case, no flow files were removed. The ISCX-VPN-App dataset was used for application classification across 13 distinct classes. Meanwhile, the USTC-TFC dataset focused on software identification, featuring 20 classes.

To mitigate potential dataset-level overfitting associated with specific fields, we excluded MAC addresses, IP addresses, and port information from all datasets before conducting evaluations. This approach prevents the model from relying solely on tricks, such as IP addresses, for classification, ensuring a more suitable evaluation of model performance.

We showcase the performance of five distinct pre-trained models: PERT\cite{he2020pert}, ET-BERT\cite{lin2022bert}, NetGPT\cite{meng2023netgpt}, YaTC\cite{zhao2023yet}, and Lens\cite{wang2024lens}. 
These pre-trained models showcase their superiority over non-pre-trained counterparts, such as K-fingerprinting\cite{hayes2016k}, FS-Net\cite{liu2019fs}, FlowPrint\cite{van2020flowprint}, and TSCRNN\cite{lin2021tscrnn}.

The Macro F1-Scores of the aforementioned five pre-trained models and our model's results are displayed in Table \ref{tab:fine-tune}. We observe that our Macro F1-Score outperforms others across most datasets, showing an average improvement of approximately 2\%. This indicates that our model achieves state-of-the-art performance in traffic classification tasks.
Another notable finding is that the overall performance of \ModelName (12k) is slightly better than \ModelName (3k), suggesting that pre-training with a longer token length may yield some benefits.

\begin{figure}
    \centering
    \includegraphics[width=0.77\linewidth]{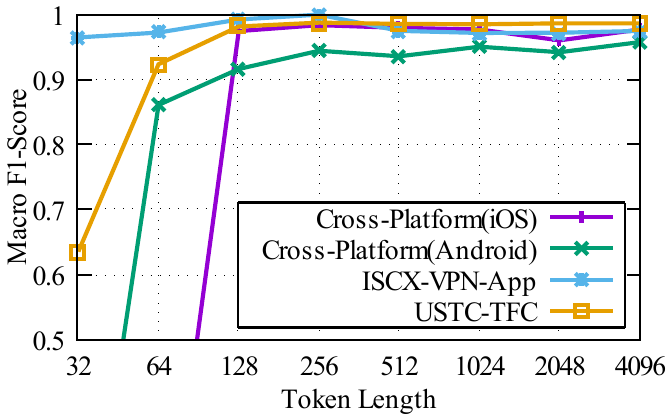}
    \caption{Variation of Macro F1-Scores with token length using \ModelName(12k) fine-tuning in classification.}
    \label{fig:token_len}
    \vspace{-5pt}
\end{figure}

In the fine-tuning process detailed in Table \ref{tab:fine-tune}, a maximum token length of 256 was established. Subsequent investigations into the impact of varying maximum token lengths on macro F1 scores, as depicted in Figure \ref{fig:token_len}, revealed a clear trend. Using \ModelName  (12k) with token lengths ranging from 32 to 4096, it was observed that classification accuracy increased sharply as the token length expanded from 32 to 128. Beyond this threshold, the improvement in F1 scores began to level off with further increases in token length. A more detailed examination of dataset-specific performances showed subtle differences; notably, on the Cross-Platform (Android) dataset, F1 scores improved consistently with longer token lengths, reaching a peak of 0.9578 for a token length of 4096. In contrast, for other datasets, optimal F1 scores were achieved at a token length of around 256, with no significant benefits from extending the token length. This analysis highlights two critical insights. Firstly, a token length of 256 generally suffices for accurate flow classification across most datasets. Secondly, increasing the token length can significantly boost classification accuracy for specific datasets, such as Cross-Platform (Android).

\begin{figure*}[ht]
    \centering
    \subfigure[HTTP Flow]{
    		\label{fig:gen:a}
     		\includegraphics[width=0.99\linewidth]{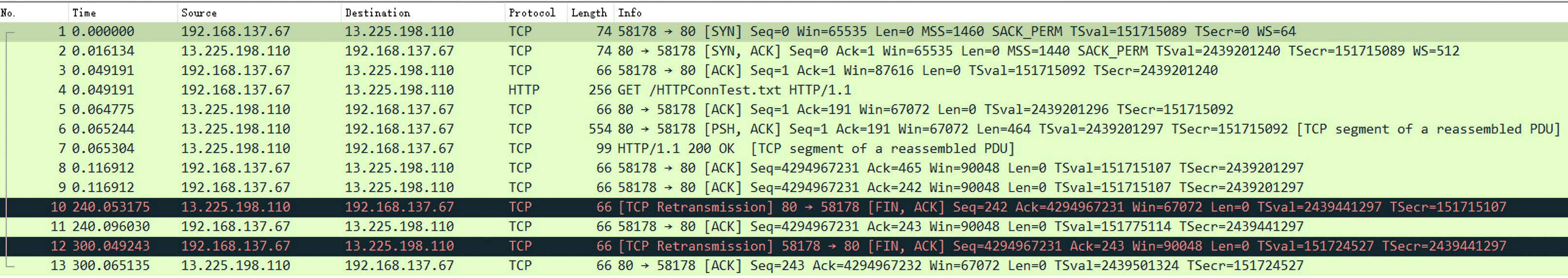}}
    \subfigure[DNS Flow]{
    		\label{fig:gen:b}
     		\includegraphics[width=0.99\linewidth]{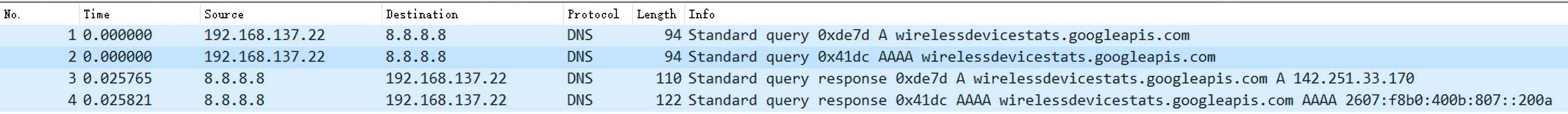}}
    \subfigure[TLS Flow]{
    		\label{fig:gen:c}
     		\includegraphics[width=0.99\linewidth]{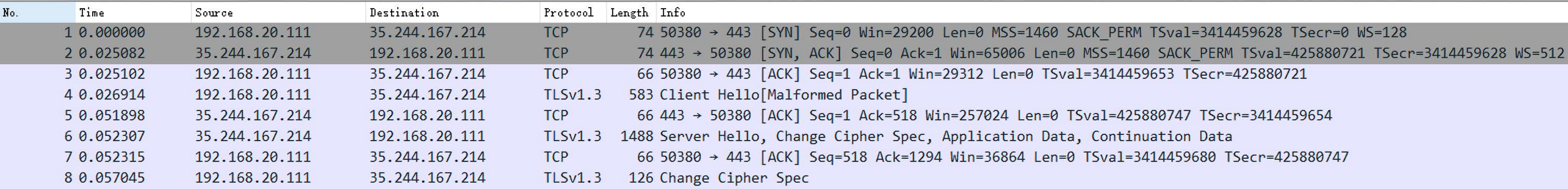}}
    \vspace{-10pt}
    \caption{{The flows generated by \ModelName(12k).}}
    \label{fig:gen}
    \vspace{-5pt}
\end{figure*}

\subsection{Generation Evaluation}
\label{subsection:gen}

\begin{figure*}[htbp]
	\centering
	\subfigure[sport]{\includegraphics[width=0.16\linewidth]{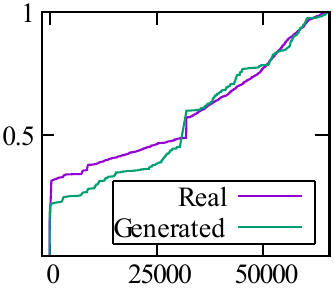}}
    \hfill
	\subfigure[dport]{\includegraphics[width=0.16\linewidth]{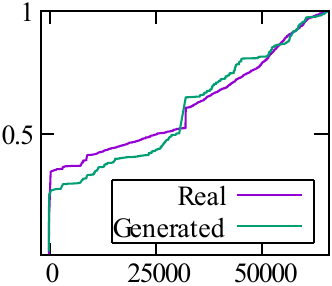}}
    \hfill
	\subfigure[src address]{\includegraphics[width=0.16\linewidth]{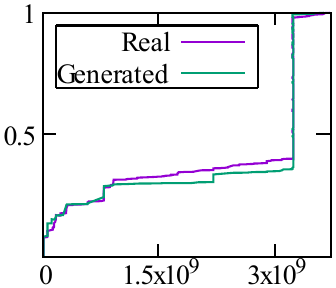}}
    \hfill
	\subfigure[dst address]{\includegraphics[width=0.16\linewidth]{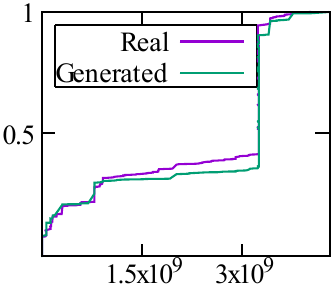}}
    \hfill
	\subfigure[packet length]{\includegraphics[width=0.16\linewidth]{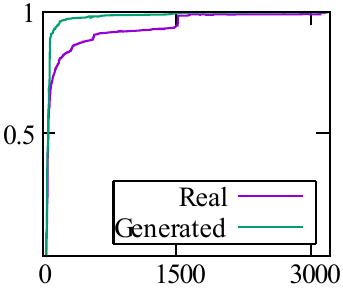}}
    \hfill
	\subfigure[TTL]{\includegraphics[width=0.16\linewidth]{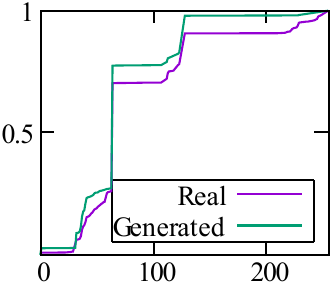}}
	\vspace{-5pt}
	\caption{ CDF plots of packet headers generated by \ModelName(12k).}
	\vspace{-2pt}
	\label{fig:packet_feature}
\end{figure*}

\begin{figure*}[htbp]
	\centering
	\subfigure[feature 1]{\includegraphics[width=0.16\linewidth]{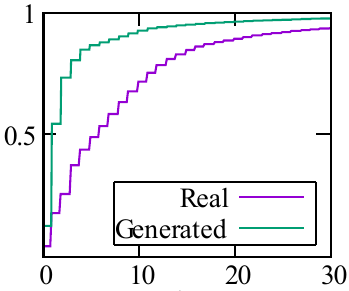}}
    \hfill
	\subfigure[feature 2]{\includegraphics[width=0.16\linewidth]{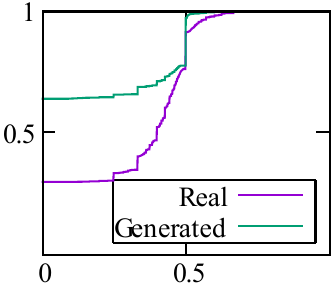}}
    \hfill
	\subfigure[feature 3]{\includegraphics[width=0.16\linewidth]{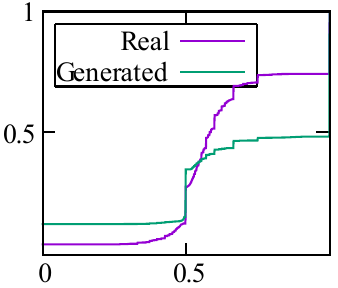}}
    \hfill
	\subfigure[feature 4]{\includegraphics[width=0.16\linewidth]{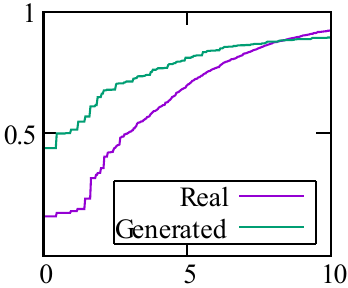}}
    \hfill
	\subfigure[feature 5]{\includegraphics[width=0.16\linewidth]{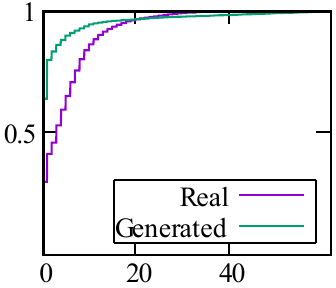}}
    \hfill
	\subfigure[feature 6]{\includegraphics[width=0.16\linewidth]{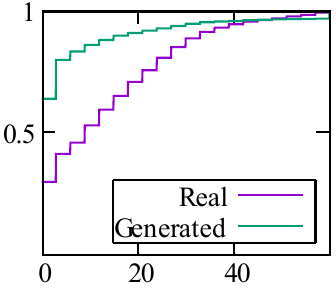}}
	\vspace{-5pt}
	\caption{ CDF plots of flow features generated by \ModelName(12k).}
	\vspace{-2pt}
	\label{fig:flow_feature}
\end{figure*}

\begin{table*}[htbp]
	\footnotesize
	\caption{Traffic Generation Performance Comparison on Packet-level JSD.}
		\vspace{0pt}
		\label{tab:JSD1} 
	\centering
	\begin{tabular}{|c|c|c|c|c|c|c|c|}
		\hline
	    \multicolumn{1}{|c|}{\textbf{Method }} & sport & dport& src address & dst address & packet length & TTL & Average \\
        \hline
		\ModelName(3k)   &0.1156  & 0.1347 & 0.1689 & 0.1779 & 0.1736 & 0.2803 & 0.1752 \\
        \hline
        \ModelName(12k)  & 0.1346  & 0.1551  &0.1684  &0.2304  &0.1874 &0.0872 & 0.1605 \\	
        \hline
	\end{tabular}
	\vspace{-10pt}
\end{table*}

\begin{table*}[htbp]
	\footnotesize
	\caption{Traffic Generation Performance Comparison on Flow-level JSD.}
		\vspace{0pt}
		\label{tab:JSD2} 
	\centering
	\begin{tabular}{|c|c|c|c|c|c|c|c|}
		\hline
	    \multicolumn{1}{|c|}{\textbf{Method }} & feature 1 & feature 2& feature 3 & feature 4 & feature 5 & feature 6 & Average \\
        \hline
		\ModelName(3k)   &0.7417 & 0.2613 & 0.1161 & 0.4051 & 0.2931 & 0.2465 & 0.3440  \\
        \hline
        \ModelName(12k)   &0.4028  &0.2529  &0.1146 & 0.3043 &0.2581 & 0.1046 & 0.2396 \\	
        \hline
	\end{tabular}
	\vspace{-10pt}
\end{table*}

In Figure \ref{fig:gen}, we present three traffic flows generated by our model, utilizing HTTP, DNS, and TLS protocols. These flows are stored in pcap format and visualized using Wireshark. Our model's traffic closely mimics authentic network patterns. In Figure \ref{fig:gen:a}, we observe precise generation of packet header fields, with the requested URL \textit{/HTTPConnTest.txt} adhering to a standard format, making it indistinguishable from real traffic. Similarly, in Figure \ref{fig:gen:b}, DNS packet requests exhibit high realism, showcasing the model's efficacy in replicating genuine traffic patterns. 
Notably, the TLS flow in Figure \ref{fig:gen:c}  demonstrates commendable generation quality. However, upon meticulous analysis, a slight deviation from standard protocol specifications is observed as a malformed Client Hello packet. 
This anomaly indicates limitations in the model's ability to generate encrypted traffic.

% 说明NLP指标可能不太适用
In the realm of generative tasks, Natural Language Processing (NLP) has witnessed substantial advancements, accompanied by the emergence of robust evaluation metrics. 
However, assessing generated network traffic poses unique challenges that conventional NLP metrics may not adequately address. Unlike text generation, network traffic generation involves intricate patterns and structures, necessitating specialized metrics for accurate evaluation.

Common NLP metrics such as BLEU\cite{papineni2002bleu}, ROUGE\cite{lin2004rouge}, or perplexity are tailored for linguistic tasks and may not effectively capture the nuances of traffic generation. 
For instance, the intricacies involved in packet header and flow feature generation demand metrics capable of quantifying dissimilarity between probability distributions and discerning subtle differences in complex structures.

In recognition of these challenges, our evaluation framework incorporates specialized metrics and discriminative models tailored to the unique demands of network traffic generation.

\noindent
\textbf{Packet Header Divergence.}
Jensen-Shannon Divergence (JSD) is employed in this study as a key evaluation metric to quantify the dissimilarity between probability distributions. Derived from information theory, JSD offers a symmetric and continuous measure of the difference between two probability distributions, making it particularly suitable for applications such as text classification and clustering. It is computed by averaging the Kullback-Leibler Divergence (KL Divergence) between each distribution and their arithmetic mean. 
The resulting metric ranges between 0 and 1, with 0 indicating perfect similarity and 1 representing complete dissimilarity. 
NetGPT utilizes JSD as a metric to assess the packet header generation quality \cite{meng2023netgpt}, and our paper adopts this idea for evaluating the performance. 

To comprehensively evaluate packet header generation, we introduced assessments for several crucial fields, including IP addresses, ports, packet lengths, and TTL. 

Figure \ref{fig:packet_feature} illustrates the Cumulative Distribution Function (CDF) plots for generated traffic compared to real traffic, while Table \ref{tab:JSD1} presents the JSD scores for generated samples. 
Two notable observations emerge. 
Firstly, the distribution of the generated data closely aligns with that of the actual data, emphasizing \ModelName (12k)'s effectiveness with an average JSD score of 0.1605. 
Secondly, the marginal discrepancies in JSD scores among various headers of the data packets suggest consistent performance across different packet header evaluations. This indicates the model's ability to maintain coherence in diverse headers of traffic generation.

% JS 散度
\noindent
\textbf{Flow Feature Divergence.}
To further evaluate the quality of flow generation on a broader scale, we introduce a divergence analysis focused on flow features.
Specifically, we derive features from the generated flows and compare them with those of the test set using JSD. Flow features play a pivotal role in tasks related to traffic analysis, and in this paper, we leverage the feature generation approach outlined in \cite{hayes2016k}. We meticulously select the top 6 effective features for computing JSD, and the specific features are detailed in Table \ref{tab:flow_features}.

The assessment of flow generation is further elucidated through Figure \ref{fig:flow_feature}, illustrating CDF plots for flow features, and Table \ref{tab:JSD2}, which presents the corresponding JSD scores. 
A noteworthy finding is the JSD score of \ModelName (12k) for flow features, standing at 0.2396, indicating a somewhat more significant deviation than the packet header scores. 
Examining the CDF plots reveals a similarity in the curves between the generated flow features and the authentic distribution, albeit with some distinctions.

These distinctions imply that generating flows may pose a more intricate challenge than generating packet headers. One plausible explanation could be the inherent complexity of capturing long-range token dependencies. Despite these challenges, the generated flow features still resemble the actual distribution, underscoring the model's capability to capture critical flow characteristics.

Additionally, the JSD scores demonstrate that the 12k model significantly outperforms the 3k model in flow generation. This observation emphasizes the positive impact of increasing token length on enhancing the effectiveness of flow generation. The longer context provided by a higher token length allows the model to better comprehend intricate patterns and dependencies within flows, resulting in a more coherent and realistic generation of network traffic flows.

\begin{table}[tbp]
%	\vspace{-5pt}
	\caption{The six features for calculating flow feature divergence.}
\label{tab:flow_features}  
	\centering
	\begin{tabular}{cl}
	\hline
	ID   & Feature Description \\
    \hline
	1  & Number of incoming packets. \\
	2   & Number of outgoing packets as a fraction of the total number \\	
        & of packets.\\
    3   & Number of incoming packets as a fraction of the total number \\  
       & of packets.\\
    4   & Standard deviation of the outgoing packet ordering list.\\  
    5   & Number of outgoing packets.\\  
    6   & Sum of all items in the alternative concentration feature list.\\   
    \hline
	\end{tabular}
	\vspace{-10pt}
\end{table}

% 基于模型的
\noindent
\textbf{Discriminative Model Assessment.}
In addition to evaluating generative models through JSD analysis, we employ discriminative models to further assess the performance and authenticity of the generated flows. Discriminative models are instrumental in distinguishing between real and synthetic data, providing a complementary perspective to generative model evaluations.
To implement discriminative model assessment, we train a classifier on the combined dataset comprising both real and generated flows. This classifier is designed to discern subtle differences and patterns between genuine and synthetic flow data. 

If the discriminator struggles to differentiate between real and generated data effectively, it suggests that the generative model has successfully captured intricate patterns and features present in the authentic dataset. This difficulty in discrimination implies that the generated flows closely resemble the characteristics of real data, demonstrating high authenticity and realism in the generated samples.

Specifically, we adopt a traffic classifier proposed by Qu \textit{et al.}  \cite{qu2023input}. This classifier employs a hierarchical structure, capable of taking packet byte inputs, making it well-suited for the context. The hierarchical nature of the classifier allows it to analyze and discern patterns at different levels of abstraction within the data, enhancing its ability to capture intricate details in both real and generated flow data.

In the experiment, 1,000 flow samples were generated and stored as pcap files, followed by a binary classification test on an equal number of randomly selected flows from the test dataset. The \ModelName (3k) model achieved a Macro F1-Score of 0.6634(±0.0412), while the \ModelName (12k) model scored slightly higher at 0.6683(±0.0232).
These results indicate a significant challenge for the discriminative model in distinguishing between real and generated flows, suggesting a high degree of realism in the generated data.

\begin{table*}[ht]
    \footnotesize
    \caption{Comparison of Traffic Classification Macro F1-Scores with Various Linear Complexity Models.}
    \vspace{-5pt}
    \label{tab:Comparative1} 
    \centering
    \begin{tabular}{|>{\centering\arraybackslash}m{2.2cm}|>{\centering\arraybackslash}m{1.3cm}|>{\centering\arraybackslash}m{1.3cm}|>{\centering\arraybackslash}m{1.3cm}|>{\centering\arraybackslash}m{1.3cm}|>{\centering\arraybackslash}m{1.3cm}|>{\centering\arraybackslash}m{1.3cm}|>{\centering\arraybackslash}m{1.3cm}|>{\centering\arraybackslash}m{1.3cm}|}
        \hline
          \multirow{2}{*}{\textbf{Method}} & \multicolumn{2}{|c|}{\textbf{Cross-Platform(iOS) }} & \multicolumn{2}{|c|}{\textbf{Cross-Platform(Android)}} & \multicolumn{2}{|c|}{\textbf{ISCX-VPN-App}} & \multicolumn{2}{|c|}{\textbf{USTC-TFC}}  \\
        \cline{2-9}
           & AC & F1 & AC & F1  & AC & F1 & AC & F1\\
        \hline
        RWKV\cite{he2020pert}&0.9275&0.8992&0.8625&0.8269&0.9750&0.9750&\textbf{0.9950}&\textbf{0.9946}\\
        RetNet\cite{sun2023retentive}&0.9675&0.9629&0.9350&0.9190&0.9906&0.9906&0.9863&0.9856\\

        \hline
        \ModelName(3k)&\textbf{0.9844}&0.9829&\textbf{0.9540}&0.9483 &{0.9912} &{0.9912} &{0.9856} &{0.9854}\\
        \ModelName(12k)&0.9839&\textbf{0.9863}&0.9444&\textbf{0.9498}&\textbf{1.0000}&\textbf{1.0000}&0.9900&0.9877\\
        \hline
    \end{tabular}
    \vspace{-5pt}
\end{table*}

\begin{table*}[ht]
	\footnotesize
	\caption{Traffic Generation Performance Comparison on Packet-level JSD with Various Linear Complexity Models.}
		\vspace{-5pt}
		\label{tab:JSD1_comp} 
	\centering
	\begin{tabular}{|c|c|c|c|c|c|c|c|}
		\hline
	    \multicolumn{1}{|c|}{\textbf{Method }} & sport & dport& src address & dst address & packet length & TTL & Average \\
        \hline
		RWKV\cite{he2020pert}&1.1841&1.2595&0.7159&1.1287&1.0192&0.6510& 0.9931 \\
        RetNet\cite{sun2023retentive}&0.9350&0.9578&1.0928&1.0264&0.9976&0.5813&0.9318 \\	
        \hline
            \ModelName (3k)   &\textbf{0.1156}  & \textbf{0.1347} & 0.1689 & \textbf{0.1779} & \textbf{0.1736} & 0.2803 & 0.1752 \\
        \ModelName (12k)  & 0.1346  & 0.1551  &\textbf{0.1684}  &0.2304  &0.1874 &\textbf{0.0872} & \textbf{0.1605} \\	
        \hline
	\end{tabular}
	\vspace{-10pt}
\end{table*}

\begin{table*}[ht]
	\footnotesize
	\caption{Traffic Generation Performance Comparison on Flow-level JSD with Various Linear Complexity Models. 'None' Represents Insufficient Data to Compute the Feature. }
		\vspace{-5pt}
		\label{tab:JSD2_comp} 
	\centering
	\begin{tabular}{|c|c|c|c|c|c|c|c|}
		\hline
	    \multicolumn{1}{|c|}{\textbf{Method }} & feature 1 & feature 2& feature 3 & feature 4 & feature 5 & feature 6 & Average \\
        \hline
		RWKV\cite{he2020pert}   & 0.8785&0.3938&0.6042&None&0.9361&1.0881& 0.7801\\
        RetNet\cite{sun2023retentive}&1.2641&0.411&0.5257&0.8547&0.6999&0.5044&0.7100  \\	
        \hline
        \ModelName (3k)   &0.7417 & 0.2613 & 0.1161 & 0.4051 & 0.2931 & 0.2465 & 0.3440  \\
        \ModelName (12k)   &\textbf{0.4028}  &\textbf{0.2529}  &\textbf{0.1146} & \textbf{0.3043} &\textbf{0.2581} & \textbf{0.1046} & \textbf{0.2396} \\	
        \hline
	\end{tabular}
	\vspace{-10pt}
\end{table*}

\subsection{Comparative Analysis of Linear Mechanisms}
In addition to \ModelName\ utilized in this paper, we also tested two other models that have shown success in NLP tasks with linear complexity: RWKV and RetNet. Their main mechanisms are detailed in Appendix \ref{sec: principle}.

For each model, we established a learning rate of $1\times 10^{-4}$, a token embedding dimension of 256, a maximum token length of 3k, and a total training step of 3,000,000. 
Due to distinct GPU memory demands for each model, we adjusted the batch size and depth to maximize GPU efficiency. For RetNet, we opted for a batch size of 8 and a model depth of 24. In the case of RWKV, constrained by hardware limitations, we set the batch size to 2 and the model depth to 18. 
It's worth noting that, for RetNet, we employed its chunkwise mode to economize on GPU memory, albeit with the trade-off of increased computation time.

Table \ref{tab:Comparative1} showcases the performance of these models in classification tasks, while Tables \ref{tab:JSD1_comp} and \ref{tab:JSD2_comp} illustrate their performance in traffic generation. Although they excel in NLP tasks, particularly classification, RWKV and RetNet demonstrate suboptimal performance in tasks related to traffic, especially in traffic generation. In our experiments, we observed that both models often generate packets that cannot be parsed correctly. We conjecture that this issue may arise from RWKV and RetNet's use of the exponential decay technique, which poses challenges in maintaining correlations between tokens over extended distances.

\section{{Discussion}}
\label{sec:Discussion}

The utilization of deep learning in traffic analysis has gained significant traction, owing to its inherent ability for automatic feature extraction\cite{rimmer2017automated, liu2020attention, luo2022transformer}. Despite its popularity, achieving high generalization poses a challenge due to the limited number of available samples. Self-supervised pre-training, obviating the need for labeled data, emerges as a pivotal strategy for acquiring and training large-scale traffic data. The superiority of pre-trained models over non-pre-trained counterparts underscores a promising direction for the future of traffic analysis.

An inherent limitation in existing self-supervised pre-trained models is the token length constraint, typically capped at 512. This constraint can severely impede the effectiveness of analysis when data packets exceed this limit, leading to a failure in capturing relationships between packets, especially in traffic generation tasks. In response to this challenge, our approach enhances the model using linear attention mechanisms, extending the token length to 12k. Experimental results validate the efficacy of this modification, demonstrating improved performance in both traffic classification and generation tasks.

While our model is pre-trained in an auto-regressive manner, which economically addresses classification and traffic generation tasks, it comes with minor drawbacks. For example, the lack of consideration for classification tasks during pre-training may introduce conceptual gaps. 
Adopting a multi-task training strategy could mitigate this limitation and enhance classification results. By incorporating classification tasks alongside auto-regressive learning during training, the model can develop a more comprehensive understanding of the data, potentially improving its performance across various tasks. 

Furthermore, the current model treats TCP and UDP flows as the basic units, overlooking information correlation between multiple flows. We recognize this as a future direction for improvement, exploring the integration of a multi-flow architecture with self-supervised learning to enhance overall performance potentially.

Regarding dataset composition, our current dataset primarily consists of TCP/IP data. However, the model architecture is designed to support packet analysis for diverse protocol stacks such as Bluetooth\cite{dong2020your,bezawada2021behavioral,barman2021every}, Zigbee\cite{babun2020z,shafqat2022zleaks,cheng2022fingerprint}, etc. This opens up an important avenue for future research, where expanding the dataset to encompass a broader range of protocols could further enhance the model's versatility and applicability.

\section{\textbf{Conclusion}}
\label{sec:Conclusion}

We developed a deep learning model \ModelName\ that is specifically designed for analyzing and generating network traffic. Our model combines generative pre-training with a linear attention mechanism to tackle the challenges associated with traditional approaches to network traffic studies. With a token limit of 12,032 tokens, our model significantly surpasses existing models in terms of capacity, enabling a more comprehensive analysis and generation of long traffic flows.
Our evaluation showcased our model's superiority in network traffic classification, where it consistently outperforms other models 
across various datasets. Moreover, in traffic generation, our model demonstrates a remarkable ability to mimic real network flows, with metrics such as JS divergence attesting to the high quality and realism of the generated traffic.

\ifCLASSOPTIONcaptionsoff
  \newpage
\fi

\bibliographystyle{IEEEtran}
\bibliography{ref}

\renewcommand{\appendixname}{\textbf{Appendices}}
\appendix

\subsection{Detailed Mechanism of Efficient Transformers}
\label{sec: principle}
We delve into the detailed mechanisms of several existing models designed to mitigate the quadratic complexity bottleneck of the Transformer. 

\noindent
\textbf{Vaswani's Attention.} The original Transformer uses Vaswani’s scaled dot-product self-attention \cite{vaswani2017attention}:
\begin{equation}
\begin{cases}
\text{Attention} = V' =\text{Softmax}(\frac{QK^T}{\sqrt{d_k}})V,\\
Q = XW^Q,\\
K = XW^K,\\
V = XW^V,
\end{cases}%\text是为了在数学公式中处理中文
\label{eq:vaswani_attention} 
\end{equation}
where $X$ is the input, $W^Q$, $W^K$ and $W^V$ are leachable parameters.
The primary computational cost of Vaswani's Attention arises from the attention calculation. Assuming the input sequence $X$ has a length of $N$ and each position's vector dimension is $d$, the computational time complexity of self-attention is $O(N^2d)$.
This is because, for each position, attention distributions to all other positions need to be computed, involving matrix multiplication with a complexity of $O(N^2)$.

\noindent
\textbf{Linear Attention.} In 2020, Katharopoulos \textit{et al.} expressed the self-attention mechanism as a linear dot-product of kernel feature maps and leveraging the associativity property of matrix products to reduce complexity from $O(N^2d)$ to $O(Nd^2)$\cite{katharopoulos2020transformers}.
The key idea is to rewrite the attention formula as follows,
\begin{equation}
V_i' =  \frac{\sum_{j=1}^N\text{sim}(Q_i,K_j)V_j}{\sum_{j=1}^N\text{sim}(Q_i,K_j)}.
\label{eq:Katharopoulos_attention} 
\end{equation}
The index $i$ is the $i$-th row vector of the matrix.
Next, they introduced a special kernel $\phi(x)$ to represent similarity, and \eqref{eq:Katharopoulos_attention} can be rewritten as
\begin{equation}
V_i' =  \frac{\sum_{j=1}^N\phi (Q_i^T) \phi (K_j) V_j}{\sum_{j=1}^N\phi (Q_i^T) \phi (K_j)},
\label{eq:Katharopoulos_attention} 
\end{equation}
and the equation can be further simplified to
\begin{equation}
V_i' =  \frac{\phi (Q_i^T) \sum_{j=1}^N \phi (K_j) V_j}{\phi (Q_i^T)\sum_{j=1}^N \phi (K_j)}.
\label{eq:Katharopoulos_attention} 
\end{equation}
So far, the Linear Transformer has exchanged the calculation sequence from $(QK)V$ to $Q(KV)$, thereby significantly reducing the computation and achieving linear complexity.

\noindent
\textbf{Reformer.} In 2020, Kitaev \textit{et al.} introduced a novel Transformer variant designed to significantly reduce memory usage and computational demands, named the Reformer\cite{kitaev2020reformer}. The Reformer addresses the limitations of traditional Transformer models by incorporating two key innovations: locality-sensitive hashing (LSH) for attention and reversible residual layers. The core idea is to replace the standard softmax attention mechanism with an LSH-based mechanism that clusters similar keys together, reducing the complexity of attention from quadratic to almost linear with respect to sequence length. The mathematical representation for the LSH attention is not as straightforward as RWKV's due to its algorithmic nature, but the essence is:
\begin{equation}
V' = \text{Aggregate}(\text{Hash}(Q), \text{Hash}(K), V),
\end{equation}
where $Q$, $K$, and $V$ are the query, key, and value matrices, respectively, and $\text{Hash}(\cdot)$ denotes the hashing function that groups similar items. Additionally, the Reformer employs reversible layers, allowing for the computation of gradients without storing activations for each layer, further reducing memory footprint. This innovative approach makes the Reformer particularly well-suited for processing long sequences, with a time complexity that scales linearly with sequence length, specifically $O(N \log N)$ due to the sorting step in LSH.

\noindent
\textbf{RWKV.} In 2023, Peng  \textit{et al.} introduced a new Transformer variant with a attention-like structure called RWKV\cite{peng2023rwkv}. This structure replaces the dot product with addition, and the formula is as follows:
\begin{equation}
V_i' =  \frac{\sum_{j=1}^N e^{W_{i,j} + K_j} V_j}{\sum_{j=1}^N e^{W_{i,j} + K_j}},
\label{eq:RWKV_attention} 
\end{equation}
in which
\begin{equation}
W_{i,j} =  - (j - i)W,
\label{eq:RWKV_attention_sub} 
\end{equation}
where $W \in (R_{\geq0})^d$, with $d$ the number of channels. The time complexity is $O(Nd)$. 

\noindent
\textbf{RetNet.} In 2023, Sun \textit{et al.} devised another Transformer-like model with linear complexity, named RetNet\cite{sun2023retentive}. It outperforms Linear Transformers and RWKV in the field of natural language processing. The parallel representation of retention is defined as:
\begin{equation}
\begin{cases}
\text{Retention} = V' = (QK^T \odot  D)V,\\
Q = (XW^Q) \odot \Theta,\\
K = (XW^K) \odot \overline{\Theta},\\
V = (XW^V).
\end{cases}%\text是为了在数学公式中处理中文
\label{eq:RetNet_attention} 
\end{equation}
In the expression, $\Theta$ and $\overline{\Theta}$ represent a relative position embedding\cite{sun2022length}, and $D \in R^{N \times N}$ combines causal masking and exponential decay along relative distance.
The parallel training has linear time complexity, but the space consumption is greater than Linear Transformer and RWKV.

\end{document}